\documentclass[conference]{IEEEtran}
\usepackage{times}

\usepackage[numbers]{natbib}
\usepackage{multicol}
\usepackage[bookmarks=true]{hyperref}

\usepackage[utf8]{inputenc} 
\usepackage[T1]{fontenc}    
\usepackage{url}            
\usepackage{booktabs}       
\usepackage{amsfonts}       
\usepackage{nicefrac}       
\usepackage{microtype}      
\usepackage{tabularx}
\usepackage{microtype}
\usepackage{graphicx}
\usepackage{subcaption}
\usepackage{hyperref}
\hypersetup{
    colorlinks=true,
    linkcolor=blue,
    filecolor=magenta,      
    urlcolor=cyan,
    citecolor=black,
}
\usepackage[utf8]{inputenc} 
\usepackage[T1]{fontenc}    
\usepackage{dsfont}
\usepackage{url}            
\usepackage{booktabs}       
\usepackage{amsfonts}       
\usepackage{nicefrac}       
\usepackage{color}
\usepackage[dvipsnames]{xcolor}
\usepackage{mathtools}
\usepackage{amsmath,amssymb}
\usepackage{bm}
\usepackage{siunitx}
\usepackage{wrapfig}
\usepackage{algorithm}
\usepackage[noend]{algorithmic}
\usepackage{lipsum}
\usepackage{makecell}
\usepackage{cite}
\usepackage{subcaption}

\usepackage{footnote}
\makesavenoteenv{tabular}
\makesavenoteenv{table}

\usepackage{amsmath}
\usepackage{amssymb}
\usepackage{mathtools}

\usepackage{amsmath,amsfonts,bm}

\def\eqref#1{equation~\ref{#1}}

\def\1{\bm{1}}

\DeclareMathAlphabet{\mathsfit}{\encodingdefault}{\sfdefault}{m}{sl}
\SetMathAlphabet{\mathsfit}{bold}{\encodingdefault}{\sfdefault}{bx}{n}

\DeclarePairedDelimiterX{\norm}[1]{\lVert}{\rVert}{#1}

\newcommand{\ie}{i.e., }
\newcommand{\eg}{e.g., }
\newcommand{\Skip}[1]{}

\begin{document}

\title{PATO: Policy Assisted TeleOperation for Scalable Robot Data Collection}

\author{
\authorblockN{
  Shivin Dass$^{*}$\authorrefmark{2}~~
  Karl Pertsch$^*$\authorrefmark{2}~~
  Hejia Zhang\authorrefmark{2}~~
  Youngwoon Lee\authorrefmark{3}~~
  Joseph J. Lim\authorrefmark{4}~~
  Stefanos Nikolaidis\authorrefmark{2}
}
\authorblockA{
  \authorrefmark{2}USC~~
  \authorrefmark{3}UC Berkeley~~
  \authorrefmark{4}KAIST
}
}

\maketitle

\renewcommand{\thefootnote}{\fnsymbol{footnote}}
\footnotetext[1]{Equal Contributions.}
\renewcommand*{\thefootnote}{\arabic{footnote}}

\begin{abstract}
Large-scale data is an essential component of machine learning as demonstrated in recent advances in natural language processing and computer vision research. However, collecting large-scale robotic data is much more expensive and slower as each operator can control only a single robot at a time. To make this costly data collection process efficient and scalable, we propose Policy Assisted TeleOperation (PATO), a system which automates part of the demonstration collection process using a learned assistive policy. PATO autonomously executes repetitive behaviors in data collection and asks for human input only when it is uncertain about which subtask or behavior to execute. We conduct teleoperation user studies both with a real robot and a simulated robot fleet and demonstrate that our assisted teleoperation system reduces human operators' mental load while improving data collection efficiency. Further, it enables a single operator to control multiple robots in parallel, which is a first step towards scalable robotic data collection. For code and video results, see \href{https://clvrai.com/pato}{clvrai.com/pato}.
\end{abstract}

\section{Introduction}
\label{sec:intro}

Recently, many works have shown impressive robot learning results from diverse, human-collected demonstration datasets~\citep{mandlekar2018roboturk, cabi2019, awopt2021corl, ebert2022bridge, rt12022arxiv}. They underline the importance of scalable robot data collection. Yet, such human demonstration collection through ``teleoperation'' is tedious and costly: tasks need to be demonstrated repeatedly and each operator can control only a single robot at a time. Research in teleoperation has focused on exploring different interfaces, such as VR controllers~\citep{zhang2018deep} and smart phones~\citep{mandlekar2018roboturk}, but does not address the aforementioned bottlenecks to scaling data collection. Thus, current teleoperation systems are badly equipped to deliver the scalability required by modern robot learning pipelines.

Our goal is to improve the scalability of robotic data collection by providing assistance to the human operator during teleoperation. We take inspiration from other fields of machine learning, such as semantic segmentation, where costly labeling processes have been substantially accelerated by providing human annotators with learned assistance systems, \eg in the form of rough segmentation estimates, that drastically reduce the labeling burden~\citep{castrejon2017annotating,acuna2018efficient}. 

Similarly, we propose to train an assistive \emph{policy} that can automate control of repeatedly demonstrated behaviors and ask for user teleoperation only when facing a novel situation or when unsure which behavior to execute. Thereby, we aim to reduce the mental load of the human operator and enable scalable teleoperation by allowing a single operator to perform data collection with multiple robots in parallel.

In order to build an assistive system for robotic data collection, we need to solve two key challenges: (1)~we need to learn assistive policies from diverse human-collected data, which is known to be challenging~\citep{mandlekar2021matters}, and (2)~we need to learn when to ask for operator input while keeping such interventions at a minimum. To address these challenges, we propose to (1)~use a powerful hierarchical policy architecture that can effectively learn from diverse multi-modal human data, and (2)~train a stochastic policy and use its output distribution to estimate its uncertainty about how to act in the current scene and which task to pursue. We then use this estimate to elicit operator input only if the assistive policy is uncertain about how to proceed.

\begin{figure}[t]
  \centering
  \includegraphics[width=1.0\linewidth]{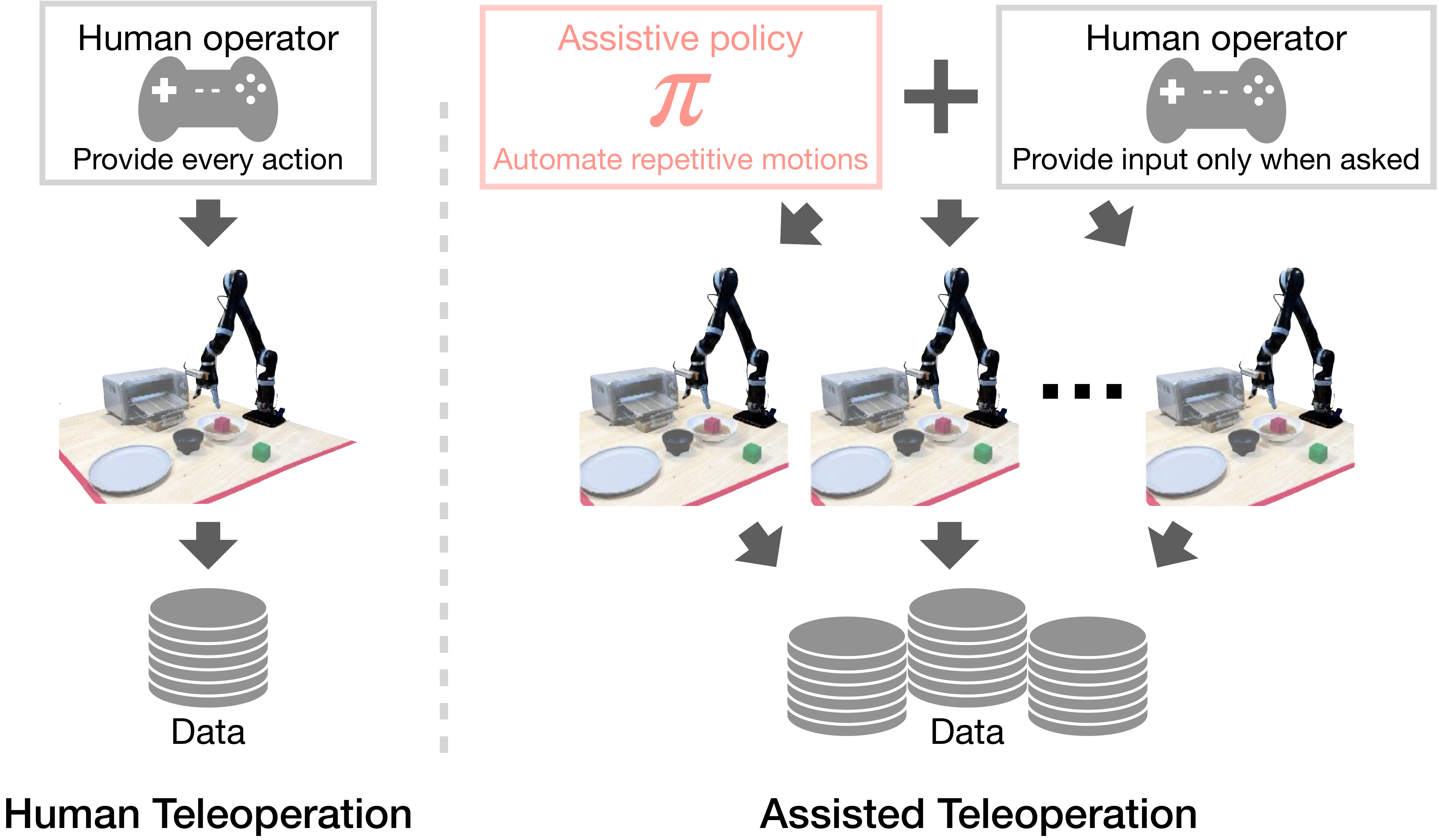}
  \caption{\textbf{Policy Assisted TeleOperation} (PATO) enables large-scale data collection by minimizing human operator inputs and mental efforts with an assistive policy, which autonomously performs repetitive subtasks. This allows a human operator to simultaneously manage multiple robots.}
  \label{fig:teaser}
\end{figure}

The main contribution of this paper is a novel Policy Assisted TeleOperation (PATO) system, which enables scalable robotic data collection using a hierarchical assistive policy. We evaluate the effectiveness of our approach in a user study in which operators collect datasets of diverse kitchen-inspired manipulation tasks with a real robot. We find that our proposed \emph{assisted} teleoperation approach reduces operators' mental load and improves their demonstration throughput. We further demonstrate that our approach allows a single operator to manage data collection with multiple robots simultaneously in a simulated manipulation environment -- a first step towards more scalable robotic data collection.
\section{Related Work}
\label{sec:related_work}

\textbf{Robot Teleoperation.} Demonstrations have played a key role in robot learning for many decades~\citep{pomerleau1989alvinn,billard2008survey,argall2009survey}; thus, many approaches have been explored for collecting such demonstrations. While initially kinesthetic teaching was common~\citep{amor2009kinesthetic} in which a human operator directly moves the robot, more recently teleoperation has become the norm~\citep{zhang2018deep,mandlekar2018roboturk,gupta2019relay,ebert2022bridge,lee2021ikea}, since separating the human operator and the robot allows for more comfortable human control inputs and is crucial for training policies with image-based inputs. Research in teleoperation systems has focused on exploring different interfaces, like VR headsets~\citep{zhang2018deep,ebert2022bridge}, joysticks~\citep{cabi2019} and smartphones~\citep{mandlekar2018roboturk}. Yet, none of these works explores active assistance of the human operator during teleoperation. Others have investigated controlling high-DoF manipulators via low-DoF interfaces through learned embedding spaces~\citep{losey2020controlling,jeon2020shared} to allow people with disabilities to control robotic arms. In contrast, our approach trains assistive policies that automate part of the teleoperation process with the goal of enabling more scalable data collection.

\textbf{Shared and Traded Autonomy.} The idea of sharing efforts between humans and robots when solving tasks has a rich history in the human-robot interaction community~\citep{javdani2015shared, selvaggio2021autonomy, berthet2016hubot, pichler2017towards, gao2017review, nikolaidis2017human, johns2016exploring, argall2018autonomy,fontaine2020quality,fontaine2022evaluating}. The literature differentiates between \emph{shared} autonomy approaches (human and robot act concurrently, \citep{crandall2002characterizing,dragan2013policy,storms2014blending}) and \emph{traded} autonomy approaches like ours (human and robot alternate control, \citep{kortenkamp1997traded,phillips2016autonomy,oh2021system}). Such approaches typically rely on a pre-defined set of goals and aim to infer the intent of the human operator to optimally assist them. Crucially, in the context of data collection, we cannot assume that all goals are known a priori, since a core goal of data collection is to collect previously unseen behaviors. Thus, instead of inferring the operator's intent over a fixed goal set, we leverage the model's estimate over its own uncertainty to determine when to assist and when to rely on operator input.

\textbf{Interactive Human Robot Learning.} In the field of robot learning, many approaches have explored leveraging human input in the learning loop and focused on different ways to decide when to leverage such input. Based on the DAgger algorithm~\citep{ross2011dagger}, many works have investigated having the human themselves decide when to intervene~\citep{kelly2019hg}, using ensemble-based support estimates~\citep{menda2019ensembledagger}, using discrepancies between model output and human inputs~\citep{zhang2016query,hoque2021lazydagger}, or risk estimates based on predicted future returns~\citep{hoque2021thriftydagger}. Yet, all these approaches focus on training a policy for a single task, not on collecting a diverse dataset. Thus, they are not designed to learn from multi-modal datasets or estimate uncertainty about the desired task. We show in our user study that these are crucial for enabling scalable robot data collection.

\textbf{Assisted Robot Data Collection.} \citet{clever2021assistive} aims to assist in robot demonstration collection via a learned policy. They visualize the projected trajectory of the assistive policy to enable the human operator to intervene if necessary. However, they focus on collection of single-task, short-horizon demonstrations and require the operator to constantly monitor the robot to decide when to intervene. In contrast, our system can collect diverse, multi-task datasets and learn when to ask the user for input, enabling more scalable data collection, \eg with multiple robots in parallel.
\section{Approach}
\label{sec:approach}

In this paper, we aim to improve the scalability of robotic data collection by learning an assistive policy to automate part of teleoperation when possible (in Section~\ref{sec:policy_learning}) and ask for user inputs only when necessary (in Section~\ref{sec:requesting_user_input}).

\subsection{Problem Formulation}
\label{sec:problem_formulation}

To enable scalable data collection of a dataset $\mathcal{D}$, we propose to automate control of repetitive behaviors using an \emph{assistive policy}. The assistive policy controls the robot and minimize required human inputs, which allows the human operator to divert attention away from the robot over contiguous intervals, \eg to attend to other robotic agents collecting data in parallel. The assistive policy can be defined as $\pi(a \vert s)$, which produces actions $a$, \eg robot end-effector displacements, given states $s$, \eg raw RGB images.

To train the assistive policy $\pi$, we assume access to a pre-collected dataset $\mathcal{D}_\text{pre}$ of diverse agent experience, \eg from scripted policies, previously collected data on different tasks or human play~\citep{lynch2020learning}. Crucially, we explicitly require our approach to handle scenarios in which the newly collected dataset $\mathcal{D}$ contains behaviors that are \emph{not} present in $\mathcal{D}_\text{pre}$. Thus, it is \emph{not} possible to fully automate data collection given the pre-training dataset $\mathcal{D}_\text{pre}$.
Instead, the system needs to request human input for unseen behaviors while providing assistance for known behaviors.

\begin{figure}[t]
  \centering
  \includegraphics[width=1.0\linewidth]{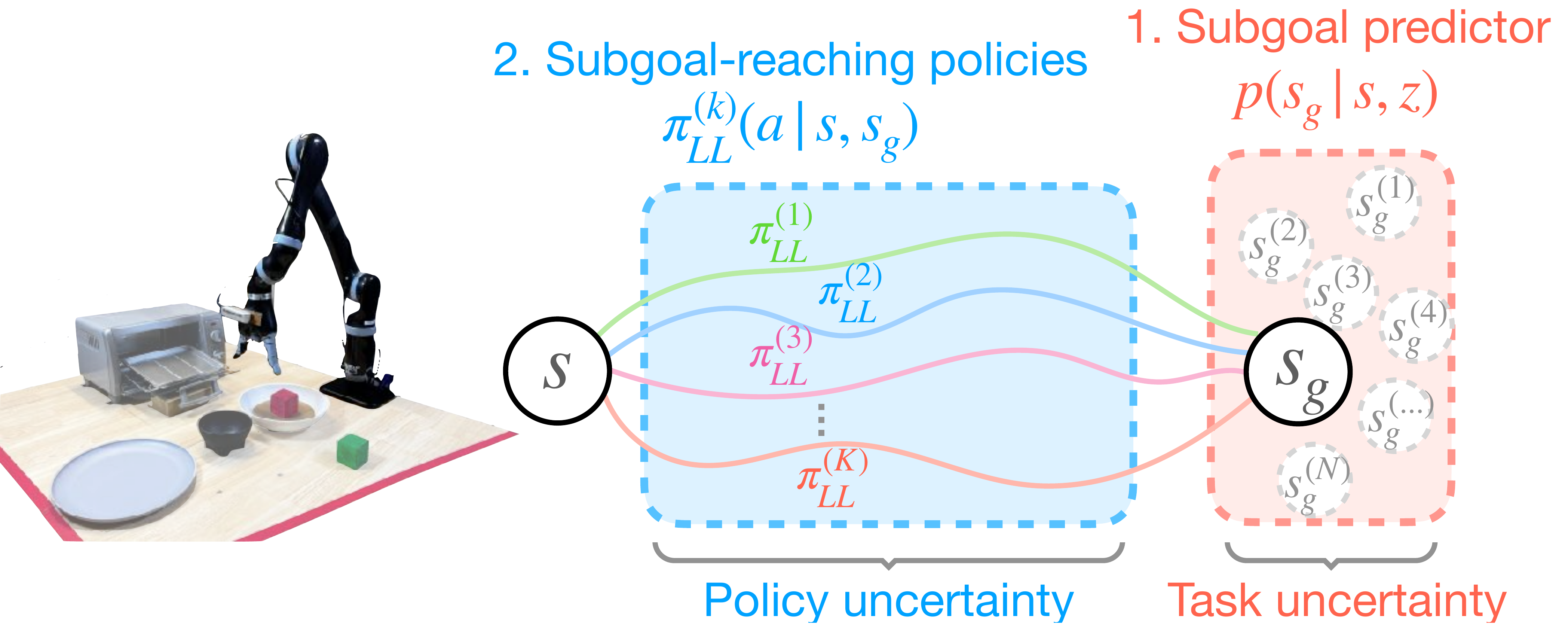}
  \caption{PATO is hierarchical: a high-level \textcolor[HTML]{FF9090}{subgoal predictor $p(s_g \vert s, z)$} and a low-level \textcolor[HTML]{3498DB}{subgoal-reaching policy $\pi_{LL}(a \vert s, s_g)$}. To decide when to follow the assistive policy, we measure uncertainty of both high-level (subgoal predictor) and low-level (subgoal-reaching policy) decisions. The task uncertainty is estimated using the subgoal predictor's variance, and the policy uncertainty is estimated as a disagreement among an ensemble of subgoal-reaching policies.}
  \label{fig:policy_model}
\end{figure}

\subsection{Learning Assistive Policies from Multi-Modal Data}
\label{sec:policy_learning}

Learning the assistive policy $\pi(a \vert s)$ from the diverse, multi-task data $\mathcal{D}_\text{pre}$ is challenging since the data is often highly multi-modal and long-horizon to imitate~\citep{mandlekar2021matters}. Thus, our assistive policy is built up on prior work in imitation of long-horizon, multi-task human data~\citep{mandlekar2019iris,gupta2019relay,lynch2020learning}.

Our hierarchical assistive policy consists of a (high-level) subgoal predictor $p(s_g \vert s, z)$ and a (low-level) subgoal-reaching policy $\pi_\text{LL}(a_t \vert s_t, s_g)$, as illustrated in Figure~\ref{fig:policy_model}. Given a state $s$, the subgoal predictor first generates a subgoal $s_g$. Then, given the subgoal and the current state, the subgoal-reaching policy outputs an action for next $\mathcal{L}$ timesteps, leading the agent toward the subgoal.

\begin{figure}[t]
  \centering
  \includegraphics[width=1.0\linewidth]{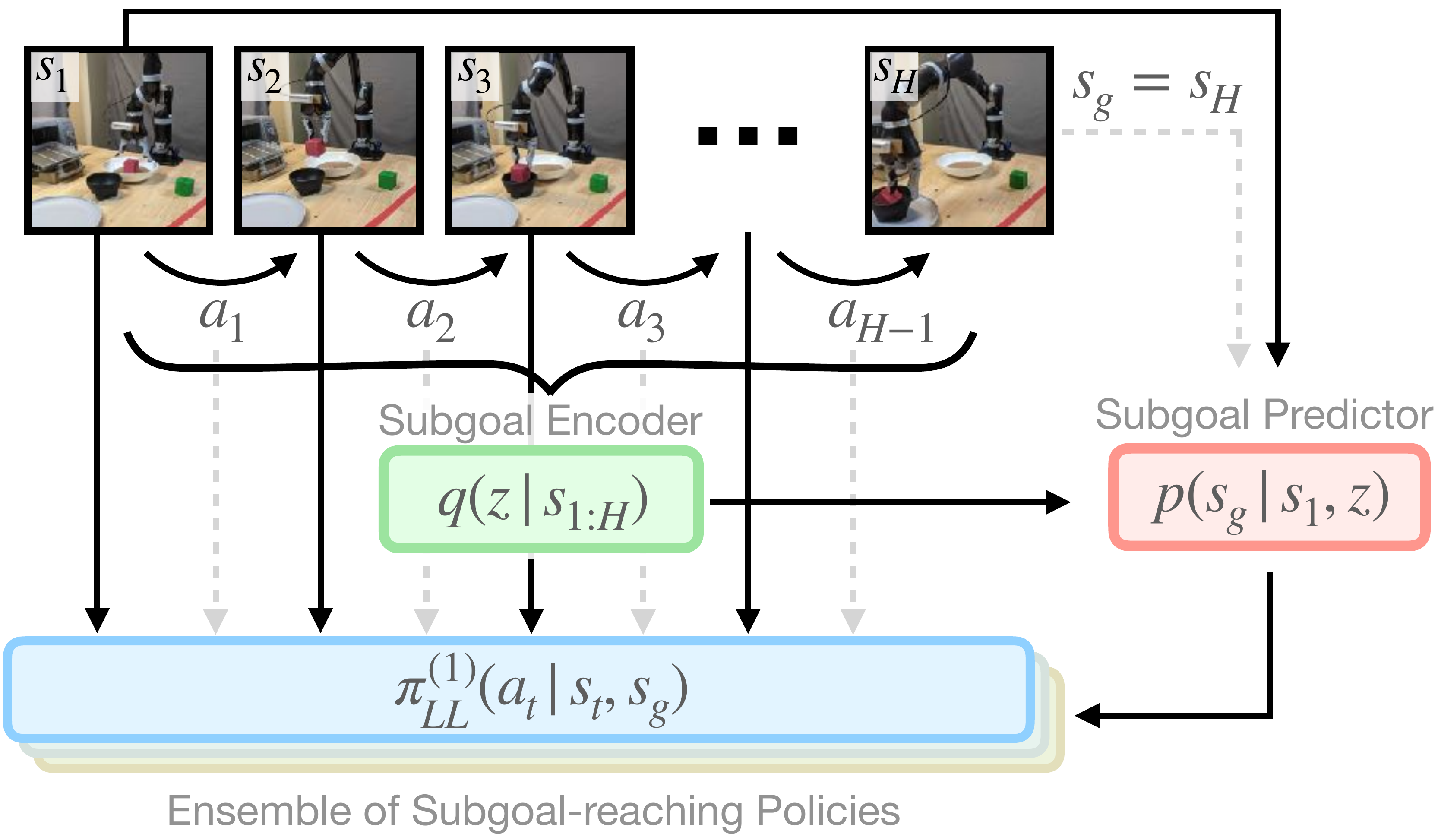}
  \caption{Our hierarchical assistive policy is trained using a pre-collected dataset $\mathcal{D}_\text{pre}$. From a sampled trajectory $(s_1, a_1, \dots, a_{\mathcal{H}-1}, s_\mathcal{H})$ of length $\mathcal{H}$, a subgoal predictor $p(s_g \vert s_1, z)$ is trained as a conditional VAE to cover a multi-modal subgoal distribution, where $s_g = s_\mathcal{H}$. Then, an ensemble of subgoal-reaching policies $\pi^{(k)}_{LL}(a_t \vert s_t, s_g)$ are trained to predict the ground truth actions. The gray dashed lines represent supervision for the prediction tasks of the subgoal predictor and subgoal-reaching policies.}
  \label{fig:model_training}
\end{figure}

To allow prediction of the \emph{full distribution} of possible subgoals in multi-modal human demonstration data, we condition the subgoal predictor on a stochastic latent variable $z$~\citep{mandlekar2019iris}. For example, in a multi-modal dataset in which the robot moves vegetables into the oven in half of the trajectories and places them on a plate in the other half, $z$ captures whether to predict a subgoal with vegetables in the oven or on the plate.

We train the subgoal predictor $p(s_g \vert s_t, z)$ as a conditional variational auto-encoder over subgoals~\citep{sohn2015learning}: given a randomly sampled starting state $s_t$ from the pre-training dataset $\mathcal{D}_\text{pre}$ and a subgoal state $s_g = s_{t+\mathcal{H}}$ $\mathcal{H}$ steps later in the trajectory, we use a learned inference network $q(z \vert s_t, s_g)$ to encode $s_t$ and $s_g$ into a latent variable $z$. We then use the subgoal predictor $p(s_g \vert s_t, z)$ to decode back to the original subgoal state $s_g$. In addition to this subgoal reconstruction objective, we apply a latent regularization loss, which shapes the distribution of the latent variable $z$ to a prior distribution $p(z)$. 

The low-level subgoal-reaching policy $\pi_{\text{LL}, \phi}(\textbf{a} \vert s, s_g)$ is trained via the behavioral cloning objective~\citep{pomerleau1989alvinn}. Since the subgoal-reaching policy needs to predict a sequence of actions given a subgoal, we opt for a recurrent policy implemented using an LSTM~\citep{hochreiter1997long}, which autoregressively predicts the actions for the next $\mathcal{L}$ steps using $s_{t}$ and $s_{g}$.

We summarize the components of our training model in Figure~\ref{fig:model_training}. Our final training objective is:
\begin{align}
    \max_{\theta, \phi, \mu} \;\; \mathbb{E}_{\substack{(s, \textbf{a}, s_g) \sim \mathcal{D}_\text{pre} \\ z \sim q(\cdot \vert s, s_g)}} \; & \underbrace{p_\theta(s_g \vert s, z)}_\text{subgoal reconstruction} + \underbrace{\pi_{\text{LL}, \phi}(\textbf{a} \vert s, s_g)}_\text{behavioral cloning}\nonumber\\
    & - \underbrace{\beta D_\text{KL}\big(q_\mu(z \vert s, s_g), p(z)\big)}_\text{latent regularization}. 
\end{align}
Here, $\textbf{a}$ represents the sequence of actions from current state until $s_g$. We use $\theta, \phi, \mu$ to denote the parameters of the subgoal predictor, goal reaching policy, and inference network, respectively. $\beta$ is a regularization weighting factor, $D_\text{KL}$ denotes the Kullback-Leibler divergence, and we use a unit Gaussian prior $p(z)$ over the latent variable. More implementation details can be found in Appendix~\ref{appendix:implementation_details}.

To execute our assistive policy $\pi(a \vert s)$, we first sample a latent variable $z$ from the unit Gaussian prior, and then generate a subgoal by passing $z$ and $s$ through the subgoal predictor $p(s_g \vert s, z)$. With the sampled subgoal $s_g$, we use the goal-reaching policy to predict an executable action $\pi_\text{LL}(a \vert s, s_g)$ for $\mathcal{L}$ timesteps. Every $\mathcal{L}$ actions, we sample a new subgoal.


\subsection{Deciding When to Request User Input}
\label{sec:requesting_user_input}

A core requirement of our approach is that it can actively ask for operator input while minimizing the number of such requests. This is crucial since it allows the operator to divert their attention to other tasks, \eg controlling other robots, while the assistive policy is controlling the robot. Thus, a key question is: when should the assistive policy ask for human inputs?

Intuitively, the assistive policy should request help when it is \emph{uncertain} about what action to take next. This can occur in two scenarios: (1)~the policy faces a situation that is not present in the training data, so it does not know which action to take, or (2)~the policy faces a seen situation, but the training trajectories contain multiple possible continuations and the policy is not sure which one to pick. The latter scenario commonly occurs during the collection of diverse datasets since trajectories for different tasks often intersect. For example, in a kitchen environment, multiple tasks could start by tossing vegetables in a pan, but then diverge into putting the pan on the stove or in the oven. During teleoperation, human inputs are required for such situations to decide which task to continue with. This is in contrast to single-task demonstration collection, \eg during DAgger training, where such uncertainty over tasks usually does not occur.

\begin{figure}[t]
  \centering
  \includegraphics[width=1.0\linewidth]{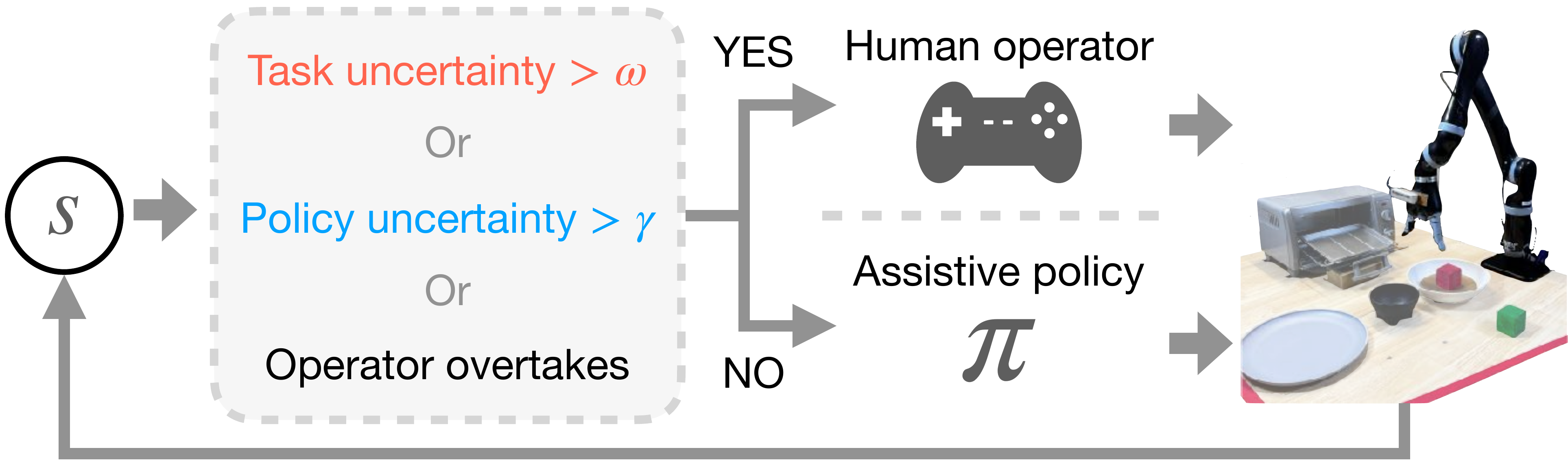}
  \caption{Our approach asks for human inputs when the assistive policy is uncertain about which subtask or action to take. If both the task uncertainty and policy uncertainty are lower than their thresholds, our assistive policy can reliably perform a subtask, reducing the workload of the human operator.}
  \label{fig:control_flow}
\end{figure}

Specifically, we let the assistive policy to ask for user input when it has a high uncertainty due to (1)~an out-of-distribution state or (2)~a multi-modal action distribution.
Our hierarchical model allows us to separately estimate both classes of uncertainty. 

First, to estimate whether a given state is unseen, we follow prior work on out-of-distribution detection~\citep{lakshminarayanan2017simple,menda2019ensembledagger} and train an ensemble of $K$ low-level subgoal-reaching policies $\pi_\text{LL}^{(1)}(a^{(1)} \vert s), \dots, \pi_\text{LL}^{(K)}(a^{(K)} \vert s)$, all on the same data $\mathcal{D}_\text{pre}$ but with different initializations and batch ordering. Then, the disagreement $D(a^{(1)}, \dots, a^{(K)})$ between the actions predicted by this ensemble of policies, i.e., the mean of the variance of actions in each dimension, can be used to approximate how out-of-distribution a state is, which we call \textbf{policy uncertainty}.
The states with high policy uncertainty (i.e. high disagreement~\citep{lakshminarayanan2017simple}) can be considered as unseen states. When the assistive policy encounters unseen states, it requests a user for deciding following actions.

Even if we determine that a state is seen in the training data, we further identify whether there are multiple possible task options in the current state before proceeding to follow the assistive policy.
To estimate the \textbf{task uncertainty}, the assistive policy's uncertainty about the task, we propose to compute the inter-subgoal variance $Var(s_g^{(1)}, \dots, s_g^{(N)})$, where the subgoals are sampled from the subgoal predictor $p_\theta(s_g^{(i)} \vert s, z^{(i)})$. If there is only one task to perform from a given state, the sampled subgoals will be similar to each other and thus, the variance of the sampled subgoals will be nearly $0$. On the other hand, in cases with multiple possible task continuations, this variance will be high since the distribution of possible subgoals will widen. If the task uncertainty is high, the policy should also stop and ask a user to choose which task to perform next.

In summary, we leverage both policy and task uncertainty estimates to decide on whether the assistive policy should continue controlling the robot or whether it should stop and ask for human input. We found a simple thresholding scheme sufficient, with threshold parameters $\gamma, \omega$ for the policy uncertainty (i.e. ensemble disagreement) and task uncertainty (i.e. subgoal variance), respectively. Future work can investigate more advanced mixing schemes, \eg with auto-tuned thresholds~\citep{hoque2021thriftydagger} or hysteresis between switching from robot to human and back~\citep{hoque2021lazydagger}. We also include a human override $H_t$ that allows the human to actively take control at any point during the teleoperation, \eg in order to demonstrate a new task from a seen state, in which case the assistive policy would not ask for human input automatically. By combining the policy uncertainty, task uncertainty, and human override, we decide to \emph{continue} executing the assistive policy if:
\begin{align}
    Assist = \;&\neg\underbrace{(D(a^{(1)}, \dots, a^{(K)}) > \gamma)}_\text{unseen state} \\
    \wedge\;&\neg\underbrace{(Var(s_g^{(1)}, \dots, s_g^{(N)}) > \omega)}_\text{uncertain task} \;\wedge\; \neg\underbrace{H_t}_\text{human override}. \nonumber
\end{align}

In our experiments, we compute the values of $\gamma$ and $\omega$ by plotting the policy uncertainty and task uncertainty over about $5$ trajectories in the pre-training dataset, similar to the plots in Figure~\ref{fig:qualitative_results}. We then choose the thresholds that correctly differentiate between the high and low uncertainty regions with the highest margin.

\vspace{1em}

\section{Experiments}
\label{sec:experiments}

This paper proposes PATO, an assistive policy for teleoperation that reduces user's mental load during data collection as well as enable one user to handle multiple robots simultaneously. 
In this section, we aim to answer the following questions: (1)~Does PATO reduce the mental load of human operators? (2)~Does it allow the operators to divert their attention to other tasks during teleoperation? (3)~Can PATO scale robotic data collection by allowing a single operator to teleoperate multiple robotic systems in parallel? To answer these questions, we perform two user studies in which (1)~users teleoperate a real robot arm to perform diverse manipulation tasks (Section~\ref{sec:robot_setup}), and (2)~users teleoperate \emph{multiple} simulated robotic arms simultaneously (Section~\ref{sec:multi_robot_experiment}). 
For implementation details of our policy architecture and used training hyperparameters, see Appendix, Section~\ref{appendix:implementation_details}.

\begin{figure}[t]
  \centering
  \includegraphics[width=1.0\linewidth]{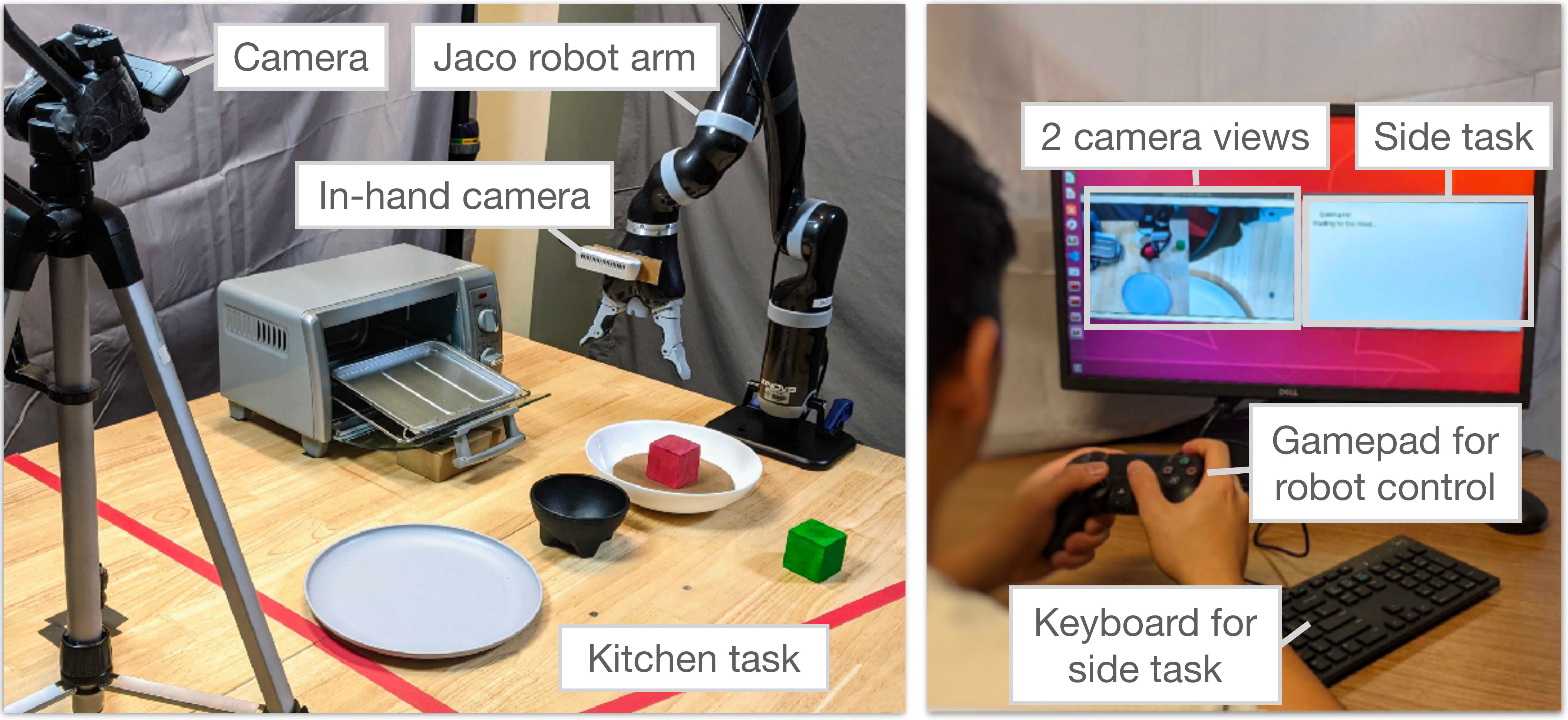}
  \caption{User study setup. \textbf{(left)} A Kinova Jaco arm, front-view and in-hand cameras, and objects for kitchen-inspired tasks are placed on the workspace. \textbf{(right)} A human operator can watch a monitor, which shows either the camera inputs or a side task. The operator uses a gamepad to control the robot, and uses a keyboard to solve the side task.}
  \label{fig:real_robot_setup}
\end{figure}

\begin{figure}[t]
    \centering
    \begin{subfigure}[t]{\linewidth}
        \centering
        \includegraphics[width=\linewidth]{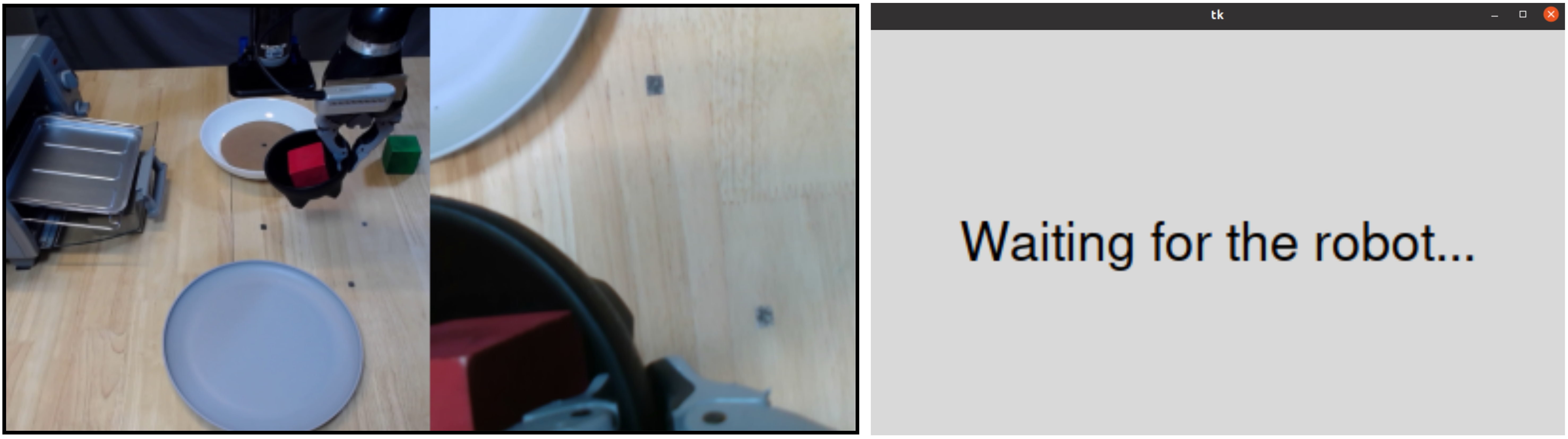}
        \caption{Side task is hidden when participant operates robot}
        \label{fig:robot_view}
    \end{subfigure}
    \par\medskip
    \begin{subfigure}[b]{\linewidth}
        \centering
        \includegraphics[width=\linewidth]{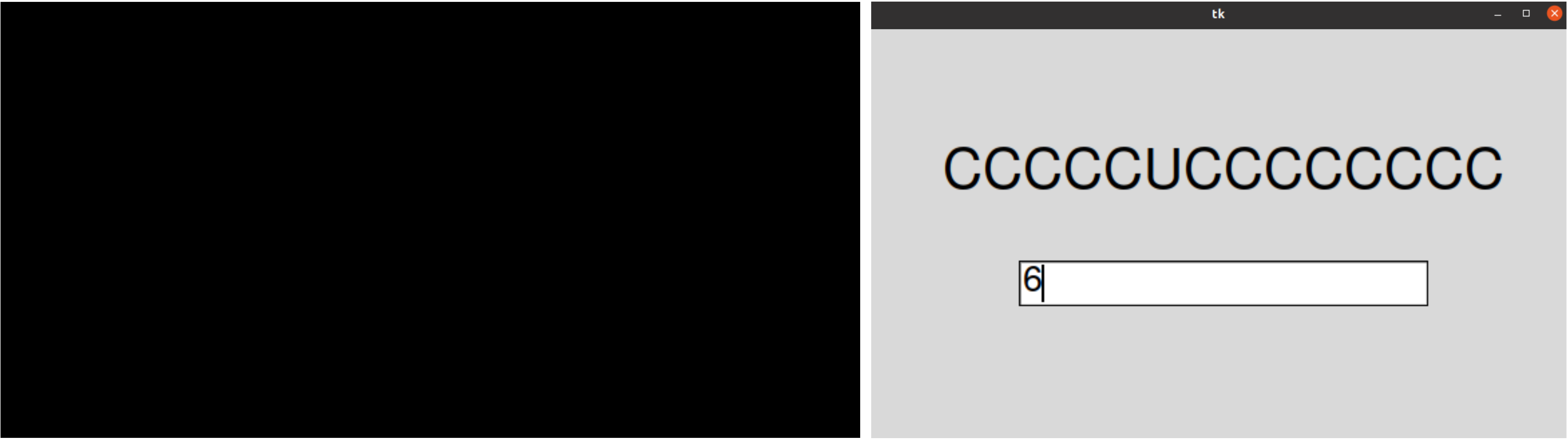}
        \caption{Robot view is hidden when participant performs the side task}
        \label{fig:side_task}
    \end{subfigure}
    \caption{Participant's screen. (a)~When a user teleoperates the robot, the side task is not visible. (b)~Similarly, while performing the side task, the robot observation screen is hided. Users can switch between robot teleoperation and side task by pressing a button on a controller.}
    \label{fig:study_example}
\end{figure}

\begin{figure*}[t]
  \centering
  \includegraphics[width=\linewidth]{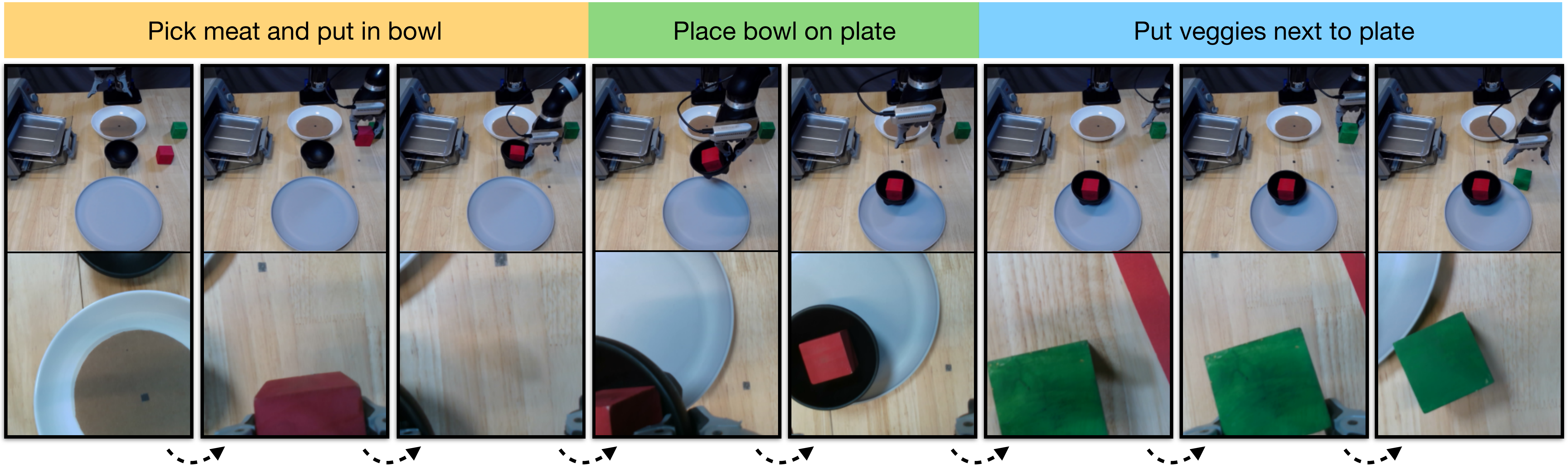}
  \caption{Visualization of the front view (top) and wrist view (bottom) of a (serve meat, serve vegetables) task trajectory in the real-world kitchen environment.}
  \label{fig:trajectory_visualization}
\end{figure*}

\subsection{Reducing Mental Load during Data Collection}
\label{sec:robot_setup}

To evaluate the effectiveness of PATO, we conduct a user study ($N = 16$) in which users teleoperate a Kinova Jaco 2 robot arm to collect diverse robot manipulation demonstrations for kitchen-inspired long-horizon tasks, \eg ``place ingredients in bowl and place bowl in oven''.

\textbf{Real-World Kitchen Environment.}\quad 
As illustrated in Figure~\ref{fig:real_robot_setup}, we evaluate our algorithm and the baselines in a real-world kitchen environment with a Kinova~Jaco~2 robot arm.
The observations are $224 \times 224 \times 3$ cropped RGB images recorded from a Logitech Pro~Webcam~C920 for front-view and an Intel RealSense~D435 for wrist-view (see Figure~\ref{fig:trajectory_visualization}). In addition to images, the neural net policies have access to robot end-effector position, velocity, and gripper state. We use a Dual~Shock~4~gamepad to control the robot's end-effector position. The action space for all policies is the delta in end-effector position and the gripper opening / closing commands. The actions are communicated at a rate of $10$Hz and translated into desired joint poses using the inverse kinematics module of PyBullet~\citep{coumans2021}. 

The environment contains $8$ long-horizon kitchen tasks composed by combining sub-tasks: \{cook meat, serve meat\}x\{cook vegetables, serve vegetables\}x\{$2$ starting locations\}. Each sub-task itself may involve the chaining of several skills. For example the sub-task \emph{cook meat} requires sequential execution of the skills \emph{pick meat from table} and \emph{put meat in oven}.

\textbf{Baseline and Prior Approach.}\quad
To train our method, PATO, we collect a pre-training dataset of 120 demonstrations. Crucially, during the user study the operators need to collect \emph{unseen} long-horizon tasks. 
We compare PATO to (1)~human teleoperation with no assistance, the current standard approach to collecting robot demonstration data and (2)~ThriftyDAgger~\citep{hoque2021thriftydagger}, the closest prior work to ours for interactive human-robot data collection. ThriftyDAgger is designed to minimize human inputs during \emph{single-task} demonstration collection by requesting human input only in \emph{critical} states where a learned value function estimates low probability for reaching the goal. We initially implemented ThriftyDAgger with a flat policy as in the original work, but found it could not learn to model the multi-modal trajectories in our training dataset, leading to poor performance. Thus, we compare to an improved version of ThriftyDAgger that uses the same hierarchical policy as ours, which we found better suited to learn from the multi-modal data.
We train all policies directly from raw RGB inputs without requiring additional state estimation.

\begin{table}[b]
\centering
\caption{Post-execution survey (Likert scales with $7$-option response format)}\label{tab:post}
\resizebox{\linewidth}{!}{%
\begin{tabular}{l}
\toprule


\textbf{Trust:}\\
Q1. I trusted the robot to do the right thing at the right time.
\vspace{1 ex}\\ 

\textbf{Robot intelligence ($\alpha = 0.95$):}\\
Q2. The robot was intelligent.\\ 
Q3. The robot perceived accurately what my goals are.\\
Q4. The robot and I worked towards mutually agreed upon goals. \\
Q5. The robot's actions were reasonable. \\
Q6. The robot did the right thing at the right time. 
\vspace{1 ex}\\ 

\textbf{Human satisfaction ($\alpha = 0.91$):}\\
Q7. I was satisfied with the robot and my performance.\\ 
Q8. The robot and I collaborated well together.\\ 
Q9. The robot was responsive to me. \\ 

\bottomrule
\end{tabular}%
}
\end{table}

\begin{figure*}[t]
  \centering
  \includegraphics[width=0.95\linewidth]{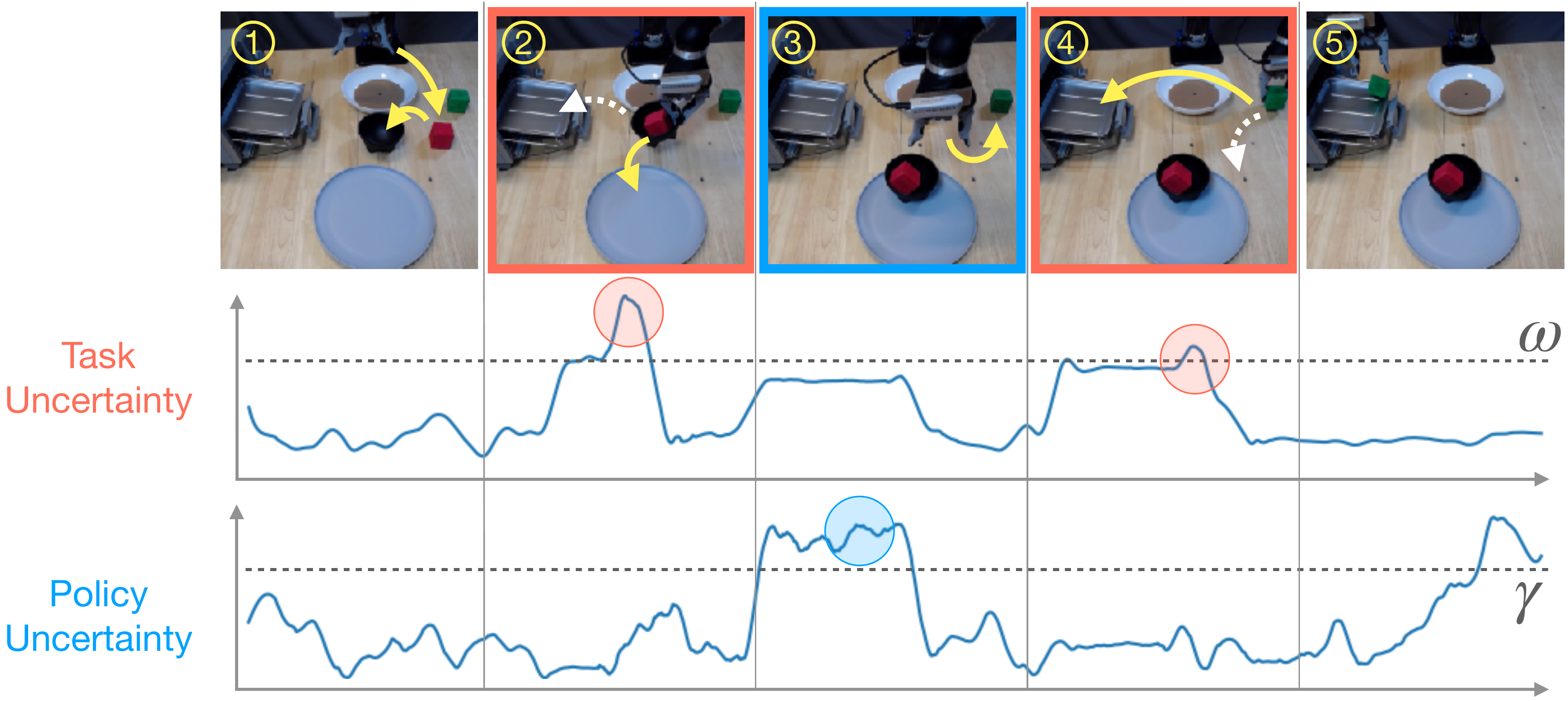}
  \caption{Visualization of PATO on a task from the real-robot user study: \textit{place red block in bowl; place bowl on plate; put green block in oven}. PATO autonomously executes familiar behaviors, but asks for user input in frames (2) and (4) to determine where to place bowl and green block (white vs. yellow arrow). In these cases, the task uncertainty surpasses the threshold $\omega$ since the subgoal predictor produces subgoals for both possible targets. Further, PATO asks for user input in frame (3) since the required transition between placing the bowl and picking up the green block was not part of its training data. Thus, the policy uncertainty estimate surpasses its threshold $\gamma$.}
  \label{fig:qualitative_results}
\end{figure*}

\textbf{User Study Design.}\quad
We recruited $16$ ($M=13$, $F=3$) participants from the graduate student population of our university. Participants were compensated with a $25$ USD gift card. The study protocol was approved by the Institutional Review Board (IRB) at our university.

All participants teleoperate the robot's end-effector via joystick and buttons on a standard gamepad controller. We give each participant $5$~minutes of unassisted training time before starting the experiment and an additional training task for each of the assisted methods (PATO and ThriftyDAgger) so that participants can familiarize themselves with the behavior of the policy. During the experiment, the participants are required to perform 3~randomly sampled long horizon tasks from a list of 8~possible tasks using each method. We counterbalance the order of the methods to guard against any sequencing effects.

The users are also asked to solve simple side tasks during teleoperation to measure their ability to divert attention and conduct other tasks. Specifically, they are shown a string of randomly selected characters, one of which is different from the others (\eg 000100), and asked to specify the index of the odd character (\eg 4 in the example above). As soon as the participant provides an answer, we present them with the next string. We leave it up to the participant how to allocate time between the robot manipulation task and the side task.

To ensure that the users can only attend to teleoperation \emph{or} the side task at a time, they perform teleoperation purely via a live-stream video of the robot setup, without being able to see the physical robot (see Figure~\ref{fig:real_robot_setup}). Crucially, they can only see \emph{either} the video stream \emph{or} the side task and can switch between the two views with a button press, as shown in Figure~\ref{fig:study_example}.

To measure the user's mental workload during teleoperation, after each teleoperation session, we administer the NASA TLX survey~\citep{hart1986nasa, hart1988development}, which measures the user's perceived workload. We aggregate the responses to obtain a single score for each user. Participants also answer a survey adopted from previous work~\citep{nikolaidis2017human}, which assesses their perception of the robot's intelligence, their satisfaction and trust in the system (Table~\ref{tab:post}).

\begin{table}[b]
\centering
\caption{Average number of completed side tasks and teleoperation time per demonstration during the real-robot teleoperation user study.}
\label{tab:unassisted_comparison}
\resizebox{1.\columnwidth}{!}{
\begin{tabular}{ccc}
    \toprule
    Approach & \multicolumn{1}{c}{Avg. \# completed side tasks} & \multicolumn{1}{c}{Avg. teleop time (sec)} \\
    \midrule
    Unassisted & $0.25$ $(\pm 0.66)$  & $109.5$ $(\pm 31.4)$  \\
    ThriftyDAgger++\footnotemark[2] & $\mathbf{13.06}$ $\mathbf{(\pm 9.63)}$  & $105.9$ $(\pm 29.5)$  \\
    PATO (ours) & $\mathbf{15.88}$ $\mathbf{(\pm 7.11)}$  &  $\mathbf{85.0}$ $\mathbf{(\pm 18.2)}$\\
    \bottomrule
\end{tabular}
}
\end{table}

\textbf{Comparison of Assisted Methods with Unassisted Baseline.}\quad
We first compare the standard unassisted teleoperation baseline to the two methods with assistance in Table~\ref{tab:unassisted_comparison}. The unassisted teleoperation baseline requires user's full attention by definition and thus, they are nearly unable to solve any side task during teleoperation. In contrast, the assisted approaches allow the users to solve additional side tasks without sacrificing teleoperation speed by diverting their attention to the side task during phases of data collection in which the policy acts autonomously.

\textbf{Comparison of PATO with ThriftyDAgger.}\quad
We then compare PATO with the prior assisted teleoperation method, ThriftyDAgger. First, we statistically evaluate the user responses to the surveys to elicit the participants' trust, satisfaction and perception of the robot's intelligence, as well as their mental workload when teleoperating with the two approaches.



During the study, participants agreed more strongly that they trusted the robot to perform the correct action at the correct time for PATO \textcolor{gray}{(Wilcoxon signed-rank test, $p = 0.001$)}\footnote[1]{Following recommended practices for Likert scales in HRI~\mbox{\citep{schrum2020four}}, we use a non-parametric Wilcoxon signed-rank test for individual Likert items, while we assume equidistant scores and use parametric ANOVA tests in multi-item scales.}. Further, they found the robot to be significantly more intelligent with the proposed method \textcolor{gray}{(repeated-measures ANOVA, $F(1, 15) = 5.14, p = 0.039$, Cronbach's $\alpha = 0.95$)} and were significantly more satisfied with their collaboration with the robot \textcolor{gray}{($F(1, 15) = 5.05, p = 0.040$, $\alpha = 0.91$)}. Finally, during the NASA TLX survey, participants showed a lower mental workload using PATO compared to the baseline \textcolor{gray}{($F(1, 15) = 5.52, p = 0.033$)}.





A key factor in these strong subjective differences between the two approaches is their ability to elicit user feedback at appropriate times: when the robot is at a decision point between two possible task continuations (see Figure~\ref{fig:qualitative_results}, frames (2) and (4)). ThriftyDAgger's risk-based objective is not sensitive to such decision points and thus, it rarely asks for user feedback. It instead executes one of the possible subtasks at random. In our study, we found that this leads to erroneous skill executions in $48$\,\% of the cases. Such errors require tedious correction by the user, deteriorating their trust in the system and their teleoperation efficiency. In contrast, as illustrated in Figure~\ref{fig:qualitative_results}, PATO leverages its estimate of task uncertainty to correctly elicit user feedback in $82$\,\% of cases, leading to higher perceived levels of trust and reduced mental load. As a result, we also find that PATO allows for more efficient data collection (see Table~\ref{tab:unassisted_comparison}) since fewer corrections are required.

\subsection{Scaling Data Collection to Multiple Robots}
\label{sec:multi_robot_experiment}

In the previous section, we showed that PATO allows users to divert their attention to other tasks during data collection. An important application of this is multi-robot teleoperation, in which a single operator performs data collection with multiple robots in parallel and periodically attends to different robots. We test this in an environment that requires simultaneous teleoperation of \emph{multiple} simulated robots.

\begin{figure}[t]
  \centering
  \includegraphics[width=\linewidth]{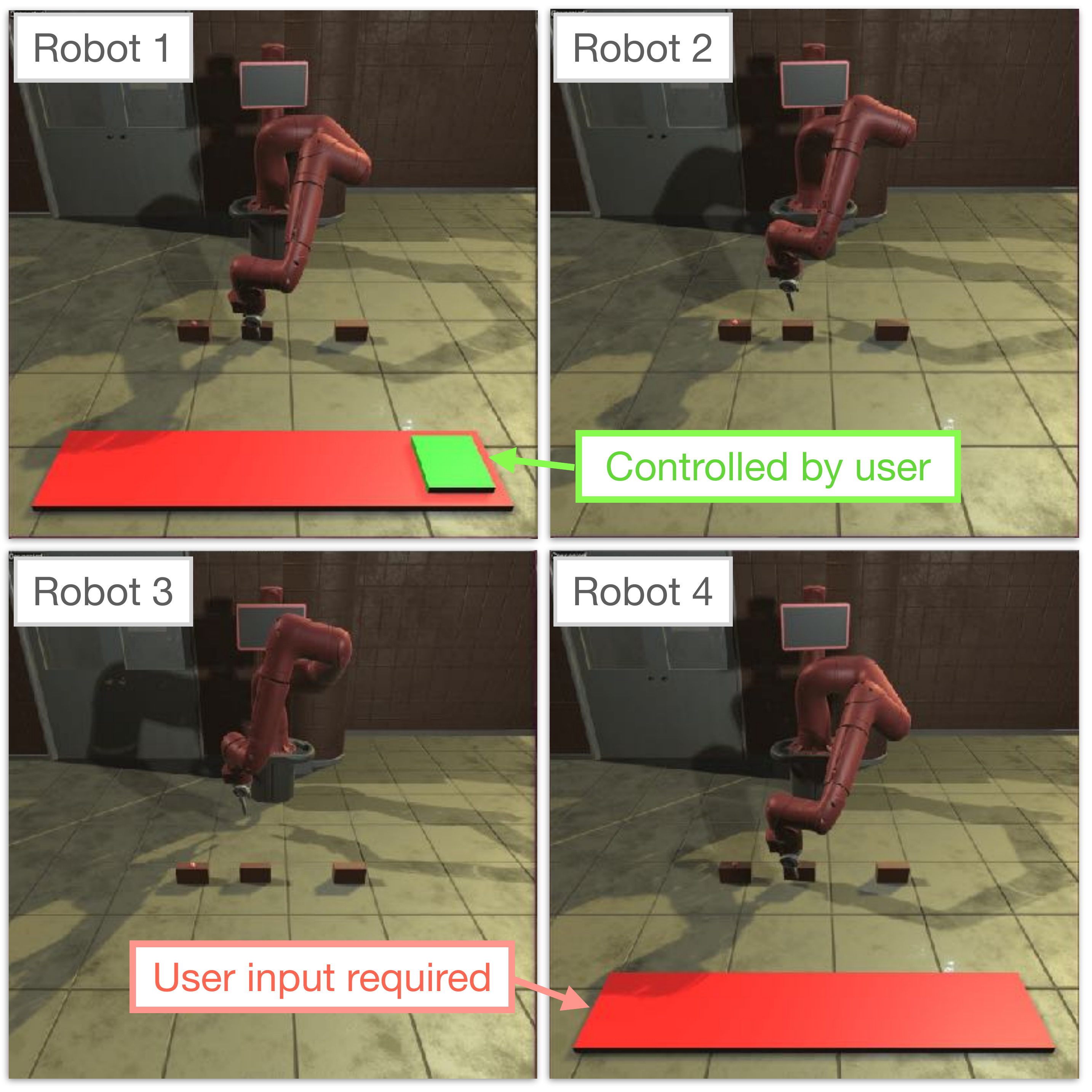}
  \caption{We use up to four simulated robots to collect demonstrations. The assistive policy asks for human input using the \textcolor{red}{red} indicator. The small \textcolor{LimeGreen}{green} indicator represents which environment is being controlled by a user.}
  \label{fig:multi_robot_setup}
\end{figure}

\textbf{Simulated IKEA Furniture Assembly Environment.}\quad 
We evaluate the scalability of data collection with assisted teleoperation on the simulated IKEA Furniture Assembly environment of \citet{lee2021ikea}, which builds on the MuJoCo physics simulator~\citep{todorov2012mujoco}. 
We use the ground truth pose and orientation measurements of the end-effector and the blocks provided by the simulator as observations for all policies. They control the robot via end-effector pose control, like in our real-world setup. We use multiple instances of this environment to simulate data collection on multiple robots, as illustrated in Figure~\ref{fig:multi_robot_setup}. We use a Dual~Shock~4~gamepad to control each robot's end-effector position, gripper open / close state and to allow switching between controlling the different robots with a button press. 

To pre-train our assistive policy, we collect 60 demonstrations for a block stacking task in which the middle block needs to be stacked on either the left block or the right block. During teleoperation time we only accept stacking on the left block as a successful trajectory. Thus, a useful assistive policy needs to elicit user input on which block to stack the middle block on, left or right, and the user needs to point the policy to the left. In this way we can measure both, whether the system asks for user input at the correct times and whether it allows users to give meaningful input to achieve their desired behavior, even when interacting with multiple robots in parallel.

\textbf{User Study Design.}\quad 
We give each participant ($N=10$) $1$~minute of training time with each of the unassisted and assisted teleoperation approaches in a single-robot version of the environment to allow them to familiarize themselves with the task and the teleoperation system. During the experiment, the participants are asked to collect as much data as possible while controlling multiple robots simultaneously~($1$, $2$ or $4$). Each participant completes data collection with all three robot fleet sizes. They are given $4$~minutes for each data collection session. For $2$ or more robots, the participants can switch between which robot they want to control via a button press and the policies can request user feedback via a visual indicator, as illustrated in Figure~\ref{fig:multi_robot_setup}. 

\footnotetext[2]{ThirftyDAgger~\citep{hoque2021thriftydagger} using our improved, hierarchical policy architecture.}

\begin{figure}[t]
  \centering
  \begin{subfigure}[t]{0.452\linewidth}
    \centering
    \includegraphics[width=1.0\linewidth]{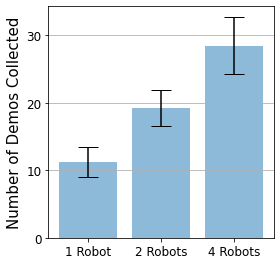}
    \caption{Scaling experiment}
    \label{fig:multi_robot_results}
  \end{subfigure}
  \hfill
  \begin{subfigure}[t]{0.531\linewidth}
    \centering
    \includegraphics[width=1.0\linewidth]{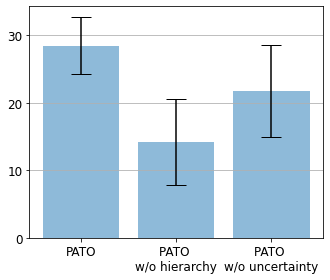}
    \caption{Ablation study}
    \label{fig:ablation_results}
  \end{subfigure}
  \caption{Average number of demonstrations collected in 4 minutes using multiple robots in simulation. \textbf{(a)}~With PATO, users can manage multiple robots simultaneously and collect more demonstrations with four robots. \textbf{(b)}~The ablated systems with four robots show inferior demonstration collection throughput.}
\end{figure}

\textbf{Multi-Robot Data Collection Results.}
We compare the number of collected demonstrations with increasing numbers of simultaneously teleoperated robots in Figure~\ref{fig:multi_robot_results}. 
Our approach enables strong scaling of data collection throughput. As expected, the scaling is not linearly proportional, \ie four robots do not lead to four times more demonstrations collected. This is because simultaneous teleoperation of a larger fleet requires more context switches between the robots, reducing the effective teleoperation time. 
Yet, PATO enables effective parallel teleoperation, leading to higher demonstration throughput with larger robot fleets. In contrast, with standard unassisted teleoperation, a single operator can only control a single robot. Thus, the demonstration throughput would not increase even if a larger fleet of robots was available.

Figure~\ref{fig:multi_robot_results} clearly shows that with the same amount of human time and effort, we can collect a larger number of demonstrations with multiple robots via our approach.
We additionally verify the quality of demonstrations collected by our parallel data collection pipeline. To this end, we train a behavioral cloning (BC) policy on the data from the multi-robot teleoperation experiment, and compare its success rates to a BC policy trained on single-robot teleoperation data. In Table~\ref{tab:multi_robot_bc_performance}, we first compare the performance of BC on data collected in $4$~minutes by unassisted single-robot teleoperation and assisted multi-robot teleoperation using PATO. The larger demonstration throughput with four robots results in a policy with higher performance using the same amount of human teleoperator effort. Additionally, we compare the BC performance on the same number of trajectories collected using single- and multi-robot teleoperation and demonstrate that the quality of data collected using PATO does not deteriorate as we increase the number of robots while only taking approximately a third of the teleoperation time. This exhibits the potential of assisted teleoperation systems to drastically increase the effectiveness of a team of operators simply by intelligently automating redundant teleoperation sequences.

\begin{table}[t]
\centering
\caption{Average success rate of behavior cloning policy trained on data from single-robot vs. multi-robot teleoperation.}
\label{tab:multi_robot_bc_performance}
\begin{tabular}{cccc}
    \toprule
    No. of robots & \multicolumn{1}{c}{Time Taken } & \multicolumn{1}{c}{\# Traj. Collected} & \multicolumn{1}{c}{Success Rate (\%)} \\
    \midrule
    $1$-robot & $4$ mins & $11$ & $34.6$ $(\pm 8.3)$  \\
    $1$-robot & \textasciitilde$10$ mins & $28$ & $\mathbf{61.0}$ $(\pm \mathbf{9.3})$  \\
    $4$-robot & $\mathbf{4}$\textbf{ mins} & $28$  & $\mathbf{56.8}$ $(\pm \mathbf{13.0})$  \\
    \bottomrule 
\end{tabular}
\end{table}

\textbf{Ablation Study.}\quad
Our approach, PATO, has two key components: the hierarchical policy (Section~\ref{sec:policy_learning}) and the uncertainty-based requesting of user input (Section~\ref{sec:requesting_user_input}). To evaluate the importance of each component, we perform ablation studies in the simulated IKEA Furniture Assembly Environment with the $4$-robot teleoperation setup described above. We compare our full method against two ablated methods: (1)~\textbf{PATO w/o hierarchy}, which trains an ensemble of \emph{flat} stochastic policies $\pi(a \vert s)$, and (2)~\textbf{PATO w/o uncertainty}, which removes the uncertainty-based requesting of user input. 
The flat policy ablation, ``PATO w/o hierarchy'', uses the ensemble disagreement and action distribution variance (instead of subgoal distribution variance in PATO) to determine policy and task uncertainty, respectively. 

We report the number of collected demonstrations per method within $4$~minutes averaged across $N = 10~\text{users}$ in Figure~\ref{fig:ablation_results}. Both ablated methods show much smaller throughputs than PATO. Specifically, we find that the flat policy in ``PATO w/o hierarchy'' is neither able to accurately model the multi-modal training data nor to accurately estimate the task uncertainty. As a result, the assistive policy does not ask for user input in critical decision states and exhibits frequent control failures, leading to reduced teleoperation throughput. Similarly, the ablation without uncertainty estimation, ``PATO w/o uncertainty'', does not elicit user input in critical states and thus, requires the user to correct it whenever it tries to complete an incorrect task. As a result, the demonstrations collected with the ablated methods are less optimal, requiring an average $398$ and $375$ steps until task completion for ``PATO w/o hierarchy'' and ``PATO w/o uncertainty'', respectively, compared to $322$ steps for our method.

\section{Conclusion}
\label{sec:conclusion}

A large-scale robot demonstration dataset is key to enabling the next breakthrough in robot learning. As a step towards large-scale robot data, we propose an efficient and scalable system for robotic data collection, which automates part of human teleoperation using a learned assistive policy and actively asks for human input in critical states. This allows a human operator to handle multiple robots simultaneously and significantly improve data collection throughput. The user study supports that our assisted teleoperation system requires infrequent inputs from users and users feel less mental burden when collecting robot demonstrations. We further show significantly improved data collection throughput of our system in the multi-robot control experiments, which leads to higher downstream policy performance for the same amount of human operator effort.

For simplicity, we assume access to pre-collected data $\mathcal{D}_\text{pre}$ to train the assistive policy. However, our assistive policy can, in theory, be learned from scratch and continuously improved as more data is collected. In this way, operators can also teach the assistive policy new skills over time and tailor its capabilities to their needs. We leave an investigation of such continually evolving assistance systems as an exciting direction for future work. Moreover, deploying our multi-robot data collection in the real world is a plausible avenue for future work.


\section*{ACKNOWLEDGMENT}
This work was supported by Institute of Information \& communications Technology Planning \& Evaluation (IITP) grant (No.2019-0-00075, Artificial Intelligence Graduate School Program, KAIST) and National Research Foundation of Korea (NRF) grant (NRF-2021H1D3A2A03103683), funded by the Korea government (MSIT).

\bibliographystyle{plainnat}
\bibliography{bibref_definitions_long,bibtex}

\begin{thebibliography}{52}
\providecommand{\natexlab}[1]{#1}
\providecommand{\url}[1]{\texttt{#1}}
\expandafter\ifx\csname urlstyle\endcsname\relax
  \providecommand{\doi}[1]{doi: #1}\else
  \providecommand{\doi}{doi: \begingroup \urlstyle{rm}\Url}\fi

\bibitem[Acuna et~al.(2018)Acuna, Ling, Kar, and Fidler]{acuna2018efficient}
David Acuna, Huan Ling, Amlan Kar, and Sanja Fidler.
\newblock Efficient interactive annotation of segmentation datasets with
  polygon-rnn++.
\newblock In \emph{CVPR}, pages 859--868, 2018.

\bibitem[Amor et~al.(2009)Amor, Berger, Vogt, and Jung]{amor2009kinesthetic}
Heni~Ben Amor, Erik Berger, David Vogt, and Bernhard Jung.
\newblock Kinesthetic bootstrapping: Teaching motor skills to humanoid robots
  through physical interaction.
\newblock In \emph{Annual conference on artificial intelligence}, 2009.

\bibitem[Argall(2018)]{argall2018autonomy}
Brenna~D Argall.
\newblock Autonomy in rehabilitation robotics: An intersection.
\newblock \emph{Annual Review of Control, Robotics, and Autonomous Systems},
  1:\penalty0 441, 2018.

\bibitem[Argall et~al.(2009)Argall, Chernova, Veloso, and
  Browning]{argall2009survey}
Brenna~D Argall, Sonia Chernova, Manuela Veloso, and Brett Browning.
\newblock A survey of robot learning from demonstration.
\newblock \emph{Robotics and autonomous systems}, 57\penalty0 (5):\penalty0
  469--483, 2009.

\bibitem[Berthet-Rayne et~al.(2016)Berthet-Rayne, Power, King, and
  Yang]{berthet2016hubot}
Pierre Berthet-Rayne, Maura Power, Hawkeye King, and Guang-Zhong Yang.
\newblock Hubot: A three state human-robot collaborative framework for bimanual
  surgical tasks based on learned models.
\newblock In \emph{ICRA}, pages 715--722. IEEE, 2016.

\bibitem[Billard et~al.(2008)Billard, Calinon, Dillmann, and
  Schaal]{billard2008survey}
Aude Billard, Sylvain Calinon, Ruediger Dillmann, and Stefan Schaal.
\newblock Survey: Robot programming by demonstration.
\newblock \emph{Handbook of robotics}, 59\penalty0 (BOOK\_CHAP), 2008.

\bibitem[Brohan et~al.(2022)Brohan, Brown, Carbajal, Chebotar, Dabis, Finn,
  Gopalakrishnan, Hausman, Herzog, Hsu, Ibarz, Ichter, Irpan, Jackson,
  Jesmonth, Joshi, Julian, Kalashnikov, Kuang, Leal, Lee, Levine, Lu, Malla,
  Manjunath, Mordatch, Nachum, Parada, Peralta, Perez, Pertsch, Quiambao, Rao,
  Ryoo, Salazar, Sanketi, Sayed, Singh, Sontakke, Stone, Tan, Tran, Vanhoucke,
  Vega, Vuong, Xia, Xiao, Xu, Xu, Yu, and Zitkovich]{rt12022arxiv}
Anthony Brohan, Noah Brown, Justice Carbajal, Yevgen Chebotar, Joseph Dabis,
  Chelsea Finn, Keerthana Gopalakrishnan, Karol Hausman, Alex Herzog, Jasmine
  Hsu, Julian Ibarz, Brian Ichter, Alex Irpan, Tomas Jackson, Sally Jesmonth,
  Nikhil Joshi, Ryan Julian, Dmitry Kalashnikov, Yuheng Kuang, Isabel Leal,
  Kuang-Huei Lee, Sergey Levine, Yao Lu, Utsav Malla, Deeksha Manjunath, Igor
  Mordatch, Ofir Nachum, Carolina Parada, Jodilyn Peralta, Emily Perez, Karl
  Pertsch, Jornell Quiambao, Kanishka Rao, Michael Ryoo, Grecia Salazar, Pannag
  Sanketi, Kevin Sayed, Jaspiar Singh, Sumedh Sontakke, Austin Stone, Clayton
  Tan, Huong Tran, Vincent Vanhoucke, Steve Vega, Quan Vuong, Fei Xia, Ted
  Xiao, Peng Xu, Sichun Xu, Tianhe Yu, and Brianna Zitkovich.
\newblock Rt-1: Robotics transformer for real-world control at scale.
\newblock In \emph{arXiv preprint arXiv:2212.06817}, 2022.

\bibitem[Cabi et~al.(2019)Cabi, Colmenarejo, Novikov, Konyushkova, Reed, Jeong,
  Zolna, Aytar, Budden, Vecerik, Sushkov, Barker, Scholz, Denil, de~Freitas,
  and Wang]{cabi2019}
Serkan Cabi, Sergio~Gomez Colmenarejo, Alexander Novikov, Ksenia Konyushkova,
  Scott Reed, Rae Jeong, Konrad Zolna, Yusuf Aytar, David Budden, Mel Vecerik,
  Oleg Sushkov, David Barker, Jonathan Scholz, Misha Denil, Nando de~Freitas,
  and Ziyu Wang.
\newblock Scaling data-driven robotics with reward sketching and batch
  reinforcement learning.
\newblock \emph{RSS}, 2019.

\bibitem[Castrejon et~al.(2017)Castrejon, Kundu, Urtasun, and
  Fidler]{castrejon2017annotating}
Lluis Castrejon, Kaustav Kundu, Raquel Urtasun, and Sanja Fidler.
\newblock Annotating object instances with a polygon-rnn.
\newblock In \emph{CVPR}, 2017.

\bibitem[Clever et~al.(2021)Clever, Handa, Mazhar, Parker, Shapira, Wan,
  Narang, Akinola, Cakmak, and Fox]{clever2021assistive}
Henry~M Clever, Ankur Handa, Hammad Mazhar, Kevin Parker, Omer Shapira, Qian
  Wan, Yashraj Narang, Iretiayo Akinola, Maya Cakmak, and Dieter Fox.
\newblock Assistive tele-op: Leveraging transformers to collect robotic task
  demonstrations.
\newblock \emph{arXiv preprint arXiv:2112.05129}, 2021.

\bibitem[Coumans and Bai(2016--2021)]{coumans2021}
Erwin Coumans and Yunfei Bai.
\newblock Pybullet, a python module for physics simulation for games, robotics
  and machine learning.
\newblock \url{http://pybullet.org}, 2016--2021.

\bibitem[Crandall and Goodrich(2002)]{crandall2002characterizing}
Jacob~W Crandall and Michael~A Goodrich.
\newblock Characterizing efficiency of human robot interaction: A case study of
  shared-control teleoperation.
\newblock In \emph{IEEE/RSJ international conference on intelligent robots and
  systems}, volume~2, pages 1290--1295. IEEE, 2002.

\bibitem[Dragan and Srinivasa(2013)]{dragan2013policy}
Anca~D Dragan and Siddhartha~S Srinivasa.
\newblock A policy-blending formalism for shared control.
\newblock \emph{The International Journal of Robotics Research}, 32\penalty0
  (7):\penalty0 790--805, 2013.

\bibitem[Ebert et~al.(2022)Ebert, Yang, Schmeckpeper, Bucher, Georgakis,
  Daniilidis, Finn, and Levine]{ebert2022bridge}
Frederik Ebert, Yanlai Yang, Karl Schmeckpeper, Bernadette Bucher, Georgios
  Georgakis, Kostas Daniilidis, Chelsea Finn, and Sergey Levine.
\newblock Bridge data: Boosting generalization of robotic skills with
  cross-domain datasets.
\newblock In \emph{RSS}, 2022.

\bibitem[Fontaine and Nikolaidis(2020)]{fontaine2020quality}
Matthew Fontaine and Stefanos Nikolaidis.
\newblock A quality diversity approach to automatically generating human-robot
  interaction scenarios in shared autonomy.
\newblock \emph{arXiv preprint arXiv:2012.04283}, 2020.

\bibitem[Fontaine and Nikolaidis(2022)]{fontaine2022evaluating}
Matthew~C Fontaine and Stefanos Nikolaidis.
\newblock Evaluating human--robot interaction algorithms in shared autonomy via
  quality diversity scenario generation.
\newblock \emph{ACM Transactions on Human-Robot Interaction (THRI)},
  11\penalty0 (3):\penalty0 1--30, 2022.

\bibitem[Gao and Chien(2017)]{gao2017review}
Yang Gao and Steve Chien.
\newblock Review on space robotics: Toward top-level science through space
  exploration.
\newblock \emph{Science Robotics}, 2\penalty0 (7):\penalty0 eaan5074, 2017.

\bibitem[Gupta et~al.(2019)Gupta, Kumar, Lynch, Levine, and
  Hausman]{gupta2019relay}
Abhishek Gupta, Vikash Kumar, Corey Lynch, Sergey Levine, and Karol Hausman.
\newblock Relay policy learning: Solving long-horizon tasks via imitation and
  reinforcement learning.
\newblock \emph{CoRL}, 2019.

\bibitem[Hart(1986)]{hart1986nasa}
Sandra~G Hart.
\newblock Nasa task load index (tlx).
\newblock 1986.

\bibitem[Hart and Staveland(1988)]{hart1988development}
Sandra~G Hart and Lowell~E Staveland.
\newblock Development of nasa-tlx (task load index): Results of empirical and
  theoretical research.
\newblock In \emph{Advances in psychology}, volume~52, pages 139--183.
  Elsevier, 1988.

\bibitem[Hochreiter and Schmidhuber(1997)]{hochreiter1997long}
Sepp Hochreiter and J{\"u}rgen Schmidhuber.
\newblock Long short-term memory.
\newblock \emph{Neural computation}, 9\penalty0 (8):\penalty0 1735--1780, 1997.

\bibitem[Hoque et~al.(2021{\natexlab{a}})Hoque, Balakrishna, Novoseller,
  Wilcox, Brown, and Goldberg]{hoque2021thriftydagger}
Ryan Hoque, Ashwin Balakrishna, Ellen Novoseller, Albert Wilcox, Daniel~S
  Brown, and Ken Goldberg.
\newblock Thriftydagger: Budget-aware novelty and risk gating for interactive
  imitation learning.
\newblock \emph{CoRL}, 2021{\natexlab{a}}.

\bibitem[Hoque et~al.(2021{\natexlab{b}})Hoque, Balakrishna, Putterman, Luo,
  Brown, Seita, Thananjeyan, Novoseller, and Goldberg]{hoque2021lazydagger}
Ryan Hoque, Ashwin Balakrishna, Carl Putterman, Michael Luo, Daniel~S Brown,
  Daniel Seita, Brijen Thananjeyan, Ellen Novoseller, and Ken Goldberg.
\newblock Lazydagger: Reducing context switching in interactive imitation
  learning.
\newblock In \emph{2021 IEEE 17th International Conference on Automation
  Science and Engineering (CASE)}, 2021{\natexlab{b}}.

\bibitem[Javdani et~al.(2015)Javdani, Srinivasa, and
  Bagnell]{javdani2015shared}
Shervin Javdani, Siddhartha~S Srinivasa, and J~Andrew Bagnell.
\newblock Shared autonomy via hindsight optimization.
\newblock In \emph{RSS}, 2015.

\bibitem[Jeon et~al.(2020)Jeon, Losey, and Sadigh]{jeon2020shared}
Hong~Jun Jeon, Dylan~P Losey, and Dorsa Sadigh.
\newblock Shared autonomy with learned latent actions.
\newblock \emph{RSS}, 2020.

\bibitem[Johns et~al.(2016)Johns, Mok, Sirkin, Gowda, Smith, Talamonti, and
  Ju]{johns2016exploring}
Mishel Johns, Brian Mok, David Sirkin, Nikhil Gowda, Catherine Smith, Walter
  Talamonti, and Wendy Ju.
\newblock Exploring shared control in automated driving.
\newblock In \emph{2016 11th ACM/IEEE International Conference on Human-Robot
  Interaction (HRI)}, pages 91--98. IEEE, 2016.

\bibitem[Kelly et~al.(2019)Kelly, Sidrane, Driggs-Campbell, and
  Kochenderfer]{kelly2019hg}
Michael Kelly, Chelsea Sidrane, Katherine Driggs-Campbell, and Mykel~J
  Kochenderfer.
\newblock Hg-dagger: Interactive imitation learning with human experts.
\newblock In \emph{ICRA}, 2019.

\bibitem[Kortenkamp et~al.(1997)Kortenkamp, Bonasso, Ryan, and
  Schreckenghost]{kortenkamp1997traded}
David Kortenkamp, R~Peter Bonasso, Dan Ryan, and Debbie Schreckenghost.
\newblock Traded control with autonomous robots as mixed initiative
  interaction.
\newblock In \emph{AAAI Symposium on Mixed Initiative Interaction}, volume~97,
  pages 89--94, 1997.

\bibitem[Lakshminarayanan et~al.(2017)Lakshminarayanan, Pritzel, and
  Blundell]{lakshminarayanan2017simple}
Balaji Lakshminarayanan, Alexander Pritzel, and Charles Blundell.
\newblock Simple and scalable predictive uncertainty estimation using deep
  ensembles.
\newblock In \emph{NIPS}, 2017.

\bibitem[Lee et~al.(2021)Lee, Hu, Yang, Yin, and Lim]{lee2021ikea}
Youngwoon Lee, Edward~S Hu, Zhengyu Yang, Alex Yin, and Joseph~J Lim.
\newblock {IKEA} furniture assembly environment for long-horizon complex
  manipulation tasks.
\newblock \emph{ICRA}, 2021.
\newblock URL \url{https://clvrai.com/furniture}.

\bibitem[Losey et~al.(2020)Losey, Srinivasan, Mandlekar, Garg, and
  Sadigh]{losey2020controlling}
Dylan~P Losey, Krishnan Srinivasan, Ajay Mandlekar, Animesh Garg, and Dorsa
  Sadigh.
\newblock Controlling assistive robots with learned latent actions.
\newblock In \emph{ICRA}, 2020.

\bibitem[Lu et~al.(2021)Lu, Hausman, Chebotar, Yan, Jang, Herzog, Xiao, Irpan,
  Khansari, Kalashnikov, and Levine]{awopt2021corl}
Yao Lu, Karol Hausman, Yevgen Chebotar, Mengyuan Yan, Eric Jang, Alexander
  Herzog, Ted Xiao, Alex Irpan, Mohi Khansari, Dmitry Kalashnikov, and Sergey
  Levine.
\newblock Aw-opt: Learning robotic skills with imitation and reinforcement at
  scale.
\newblock In \emph{CoRL}, 2021.

\bibitem[Lynch et~al.(2020)Lynch, Khansari, Xiao, Kumar, Tompson, Levine, and
  Sermanet]{lynch2020learning}
Corey Lynch, Mohi Khansari, Ted Xiao, Vikash Kumar, Jonathan Tompson, Sergey
  Levine, and Pierre Sermanet.
\newblock Learning latent plans from play.
\newblock In \emph{CoRL}, 2020.

\bibitem[Mandlekar et~al.(2018)Mandlekar, Zhu, Garg, Booher, Spero, Tung, Gao,
  Emmons, Gupta, Orbay, Savarese, and Fei-Fei]{mandlekar2018roboturk}
Ajay Mandlekar, Yuke Zhu, Animesh Garg, Jonathan Booher, Max Spero, Albert
  Tung, Julian Gao, John Emmons, Anchit Gupta, Emre Orbay, Silvio Savarese, and
  Li~Fei-Fei.
\newblock Roboturk: A crowdsourcing platform for robotic skill learning through
  imitation.
\newblock In \emph{CoRL}, 2018.

\bibitem[Mandlekar et~al.(2020{\natexlab{a}})Mandlekar, Ramos, Boots, Fei-Fei,
  Garg, and Fox]{mandlekar2019iris}
Ajay Mandlekar, Fabio Ramos, Byron Boots, Li~Fei-Fei, Animesh Garg, and Dieter
  Fox.
\newblock Iris: Implicit reinforcement without interaction at scale for
  learning control from offline robot manipulation data.
\newblock \emph{ICRA}, 2020{\natexlab{a}}.

\bibitem[Mandlekar et~al.(2020{\natexlab{b}})Mandlekar, Xu,
  Mart{\'\i}n-Mart{\'\i}n, Savarese, and Fei-Fei]{mandlekar2020gti}
Ajay Mandlekar, Danfei Xu, Roberto Mart{\'\i}n-Mart{\'\i}n, Silvio Savarese,
  and Li~Fei-Fei.
\newblock Gti: Learning to generalize across long-horizon tasks from human
  demonstrations.
\newblock In \emph{RSS}, 2020{\natexlab{b}}.

\bibitem[Mandlekar et~al.(2021)Mandlekar, Xu, Wong, Nasiriany, Wang, Kulkarni,
  Fei-Fei, Savarese, Zhu, and Mart{\'\i}n-Mart{\'\i}n]{mandlekar2021matters}
Ajay Mandlekar, Danfei Xu, Josiah Wong, Soroush Nasiriany, Chen Wang, Rohun
  Kulkarni, Li~Fei-Fei, Silvio Savarese, Yuke Zhu, and Roberto
  Mart{\'\i}n-Mart{\'\i}n.
\newblock What matters in learning from offline human demonstrations for robot
  manipulation.
\newblock \emph{CoRL}, 2021.

\bibitem[Menda et~al.(2019)Menda, Driggs-Campbell, and
  Kochenderfer]{menda2019ensembledagger}
Kunal Menda, Katherine Driggs-Campbell, and Mykel~J Kochenderfer.
\newblock Ensembledagger: A bayesian approach to safe imitation learning.
\newblock In \emph{IROS}, 2019.

\bibitem[Nair et~al.(2022)Nair, Rajeswaran, Kumar, Finn, and
  Gupta]{nair2022r3m}
Suraj Nair, Aravind Rajeswaran, Vikash Kumar, Chelsea Finn, and Abhinav Gupta.
\newblock R3m: A universal visual representation for robot manipulation.
\newblock In \emph{CoRL}, 2022.

\bibitem[Nikolaidis et~al.(2017)Nikolaidis, Zhu, Hsu, and
  Srinivasa]{nikolaidis2017human}
Stefanos Nikolaidis, Yu~Xiang Zhu, David Hsu, and Siddhartha Srinivasa.
\newblock Human-robot mutual adaptation in shared autonomy.
\newblock In \emph{2017 12th ACM/IEEE International Conference on Human-Robot
  Interaction (HRI)}, pages 294--302. IEEE, 2017.

\bibitem[Oh et~al.(2021)Oh, Sch{\"a}fer, R{\"u}ther, Toussaint, and
  Mainprice]{oh2021system}
Yoojin Oh, Tim Sch{\"a}fer, Benedikt R{\"u}ther, Marc Toussaint, and Jim
  Mainprice.
\newblock A system for traded control teleoperation of manipulation tasks using
  intent prediction from hand gestures.
\newblock In \emph{2021 30th IEEE International Conference on Robot \& Human
  Interactive Communication (RO-MAN)}, pages 503--508. IEEE, 2021.

\bibitem[Phillips-Grafflin et~al.(2016)Phillips-Grafflin, Suay, Mainprice,
  Alunni, Lofaro, Berenson, Chernova, Lindeman, and Oh]{phillips2016autonomy}
Calder Phillips-Grafflin, Halit~Bener Suay, Jim Mainprice, Nicholas Alunni,
  Daniel Lofaro, Dmitry Berenson, Sonia Chernova, Robert~W Lindeman, and Paul
  Oh.
\newblock From autonomy to cooperative traded control of humanoid manipulation
  tasks with unreliable communication: Applications to the valve-turning task
  of the darpa robotics challenge and lessons learned.
\newblock \emph{Journal of Intelligent \& Robotic Systems}, 82:\penalty0
  341--361, 2016.

\bibitem[Pichler et~al.(2017)Pichler, Akkaladevi, Ikeda, Hofmann, Plasch,
  W{\"o}gerer, and Fritz]{pichler2017towards}
Andreas Pichler, Sharath~Chandra Akkaladevi, Markus Ikeda, Michael Hofmann,
  Matthias Plasch, Christian W{\"o}gerer, and Gerald Fritz.
\newblock Towards shared autonomy for robotic tasks in manufacturing.
\newblock \emph{Procedia Manufacturing}, 11:\penalty0 72--82, 2017.

\bibitem[Pomerleau(1989)]{pomerleau1989alvinn}
Dean~A Pomerleau.
\newblock Alvinn: An autonomous land vehicle in a neural network.
\newblock In \emph{NIPS}, pages 305--313, 1989.

\bibitem[Ross et~al.(2011)Ross, Gordon, and Bagnell]{ross2011dagger}
St{\'e}phane Ross, Geoffrey Gordon, and Drew Bagnell.
\newblock A reduction of imitation learning and structured prediction to
  no-regret online learning.
\newblock In \emph{AISTATS}, pages 627--635, 2011.

\bibitem[Schrum et~al.(2020)Schrum, Johnson, Ghuy, and
  Gombolay]{schrum2020four}
Mariah~L Schrum, Michael Johnson, Muyleng Ghuy, and Matthew~C Gombolay.
\newblock Four years in review: Statistical practices of likert scales in
  human-robot interaction studies.
\newblock In \emph{Companion of the 2020 ACM/IEEE International Conference on
  Human-Robot Interaction}, pages 43--52, 2020.

\bibitem[Selvaggio et~al.(2021)Selvaggio, Cognetti, Nikolaidis, Ivaldi, and
  Siciliano]{selvaggio2021autonomy}
Mario Selvaggio, Marco Cognetti, Stefanos Nikolaidis, Serena Ivaldi, and Bruno
  Siciliano.
\newblock Autonomy in physical human-robot interaction: A brief survey.
\newblock \emph{IEEE Robotics and Automation Letters}, 2021.

\bibitem[Sohn et~al.(2015)Sohn, Lee, and Yan]{sohn2015learning}
Kihyuk Sohn, Honglak Lee, and Xinchen Yan.
\newblock Learning structured output representation using deep conditional
  generative models.
\newblock In \emph{NIPS}, 2015.

\bibitem[Storms and Tilbury(2014)]{storms2014blending}
Justin~G Storms and Dawn~M Tilbury.
\newblock Blending of human and obstacle avoidance control for a high speed
  mobile robot.
\newblock In \emph{2014 American Control Conference}, pages 3488--3493. IEEE,
  2014.

\bibitem[Todorov et~al.(2012)Todorov, Erez, and Tassa]{todorov2012mujoco}
Emanuel Todorov, Tom Erez, and Yuval Tassa.
\newblock Mujoco: A physics engine for model-based control.
\newblock In \emph{IROS}, 2012.

\bibitem[Zhang and Cho(2017)]{zhang2016query}
Jiakai Zhang and Kyunghyun Cho.
\newblock Query-efficient imitation learning for end-to-end autonomous driving.
\newblock \emph{AAAI}, 2017.

\bibitem[Zhang et~al.(2018)Zhang, McCarthy, Jow, Lee, Chen, Goldberg, and
  Abbeel]{zhang2018deep}
Tianhao Zhang, Zoe McCarthy, Owen Jow, Dennis Lee, Xi~Chen, Ken Goldberg, and
  Pieter Abbeel.
\newblock Deep imitation learning for complex manipulation tasks from virtual
  reality teleoperation.
\newblock In \emph{ICRA}, 2018.

\end{thebibliography}

\clearpage
\newpage
\appendix

\subsection{Implementation Details}
\label{appendix:implementation_details}
Here we provide a detailed overview of the model architecture and used hyperparameters (for a detailed listing of all hyperparameters, see Table~\ref{tab:hyperparam}). 


\subsubsection{Sub-goal Predictor} 
The conditional-VAE for the sub-goal predictor consists of an encoder and a decoder, both implemented with 5-layer MLPs with 128 hidden units in each layer, batch norm and LeakyReLU activations. The input to the encoder is the state and goal $\mathcal{H}$ steps in the future, where $\mathcal{H}$ is the goal horizon. For image observations in the real-world kitchen environment we first run the inputs through a pre-trained visual encoder, R3M~\citep{nair2022r3m}, to obtain a 2048-dimensional vector representation. The output of the encoder is the mean and log standard deviation of a 128-dimensional multivariate Gaussian latent space. The decoder uses the current state and a sample from the encoder's output distribution to generate a goal prediction. During inference, we generate goals with the decoder by sampling the latent vector from a standard Gaussian prior. $\mathcal{H}$ is set to 35 for the real-world kitchen environment and 7 for the IKEA Furniture Assembly environment.

To model the task uncertainty when deciding whether to request human input, we sample $N=1024$ goals and measure the variance between the end-effector positions of the robot. In case of high task uncertainty, the goals predicted by the sub-goal predictor diverge, leading to high variance. When the variance is higher than a threshold $\tau_{task}$, we query the human to disambiguate which task to perform next.

\subsubsection{Low-level Sub-goal Reaching Policy}
The inputs to the sub-goal reaching policy is the current state $s_{t}$ and the goal state $s_{t+\mathcal{H}}$ where $\mathcal{H}$ is the goal horizon used in the sub-goal predictor. The policy is implemented as a LSTM~\citep{hochreiter1997long} with 256-dimensional hidden units, which autoregressively predicts the actions for the next $\mathcal{L}$ steps using $s_{t}$ and $s_{t+\mathcal{H}}$. Notice that in prior works~\citep{mandlekar2020gti} skill horizon $\mathcal{L}$ and goal horizon $\mathcal{H}$ are set to the same value. However, in our experiments we noticed that setting a higher goal horizon helps model the task uncertainty better but a higher skill horizon would reduce the decision frequency of the policy. Hence, in this work, we decouple the two and show that the goal horizon $\mathcal{H}$ can be much greater than the skill horizon $\mathcal{L}$ without affecting the policy performance. The state and goal are processed by input MLPs separately and concatenated before being processed by the LSTM. The hidden state of the LSTM is processed by an output MLP to generate actions.

To model policy uncertainty, we use an ensemble of K=5 sub-goal reaching policies. Similar to \citet{menda2019ensembledagger}, we measure the variance between the actions predicted by each of the policies. When the state input to the policies is seen in the training data, we expect the action predictions of the ensemble policies to agree, leading to a low variance. When the state input is outside the distribution of seen training data, the policy predictions diverge, leading to a high variance. When variance is higher than a threshold $\tau_{policy}$, the policy queries human help.
\begin{table}[ht]
\centering
\caption{PATO hyperparameters. Parameters that differ between environments are marked in \textcolor{red}{red}.}
\label{tab:hyperparam}
\resizebox{1.\columnwidth}{!}{
\begin{tabular}{lp{2cm}p{2cm}}
    \toprule
    Hyperparameters & Kitchen & IKEA Assembly \\
    \midrule
    \multicolumn{3}{l}{\textbf{Sub-goal Predictor}}\\
    \midrule
    train iterations & \textcolor{red}{200} & \textcolor{red}{500} \\
    batch size & 16 & 16 \\
    learning rate & 0.001 & 0.001\\
    optimizer & Adam(0.9, 0.999) & Adam(0.9, 0.999) \\
    encoder-mlp (width x depth) & 128x5 & 128x5 \\
    decoder-mlp (width x depth) & 128x5 & 128x5 \\
    latent dimension & 128 & 128 \\
    normalization & batch & batch \\
    activation & LeakyReLU(0.2) & LeakyReLU(0.2)\\
    goal horizon ($\mathcal{H}$) & \textcolor{red}{35} & \textcolor{red}{7} \\
    loss & ELBO & ELBO \\
    \midrule
    \multicolumn{3}{l}{\textbf{Low-level Sub-goal Reaching Policy}}\\
    \midrule
    train iterations & \textcolor{red}{300} & \textcolor{red}{4000} \\
    batch size & 16 & 16 \\
    learning rate & 0.001 & 0.001\\
    optimizer & Adam(0.9, 0.999) & Adam(0.9, 0.999) \\
    input-mlp (width x depth) & 256x3 & 256x3 \\
    output-mlp (width x depth) & 256x3 & 256x3 \\
    lstm hidden dimension & 256 & 256 \\
    normalization & batch & batch \\
    activation & LeakyReLU(0.2) & LeakyReLU(0.2)\\
    no. of ensemble policies & 5 & 5 \\
    skill horizon ($\mathcal{L}$) & \textcolor{red}{15} & \textcolor{red}{7}\\
    loss - delta EEF positions & MSE & MSE \\
    loss - gripper & CE Loss & CE Loss \\
    \bottomrule
\end{tabular}
}
\end{table}

\subsubsection{Q-Function (ThriftyDAgger)} The ThriftyDAgger baseline~\citep{hoque2021thriftydagger} uses a risk measure derived from a trained Q-function for choosing when to request human input. The Q-function is modeled using a 3-layer MLP with 128 hidden units and LeakyReLU(0.2) activation. More details on training the Q-funtion and using it for risk calculation can be found in \citep{hoque2021thriftydagger}.





\end{document}